%% file: latex/main.tex
\documentclass[11pt]{article}

\usepackage[preprint]{acl}
\usepackage{CJKutf8}

\usepackage{times}
\usepackage{latexsym}
\usepackage{helvet}
\usepackage[T1]{fontenc}
\usepackage[utf8]{inputenc}
\usepackage{microtype}
\usepackage{inconsolata}

\usepackage{graphicx}
\usepackage{booktabs}
\usepackage{amsmath, amssymb}
\usepackage{enumitem}
\usepackage{multirow}
\usepackage{xcolor}
\usepackage{pifont}
\usepackage{longtable}
\usepackage{float}
\usepackage{tikz}
\usepackage{todonotes}
\usetikzlibrary{shapes.geometric, arrows.meta, positioning, calc, fit, backgrounds}
\usepackage[most]{tcolorbox}
\usepackage{placeins}

\usepackage{hyperref}
\hypersetup{
    colorlinks=true,
    linkcolor=blue,
    citecolor=blue,
    urlcolor=blue
}

\title{Beyond the Needle's Illusion: Decoupled Evaluation of Evidence Access and Use under Semantic Interference at 326M-Token Scale}

\author{
  \textbf{Tianwei Lin}\thanks{Equal contribution.}\textsuperscript{,1,2} \quad
  \textbf{Zuyi Zhou}\footnotemark[1]\textsuperscript{,1,2} \quad
  \textbf{Xinda Zhao}\textsuperscript{1,2} \quad
  \textbf{Chenke Wang}\textsuperscript{1,2} \quad
  \textbf{Xiaohong Li}\textsuperscript{1,2} \\
  \textbf{Yu Chen}\textsuperscript{1,2} \quad
  \textbf{Chuanrui Hu}\thanks{Corresponding author.}\textsuperscript{,1,2} \quad
  \textbf{Jian Pei}\footnotemark[2]\textsuperscript{,3} \quad
  \textbf{Yafeng Deng}\footnotemark[2]\textsuperscript{,1,2} \\[0.2em]
  \textsuperscript{1}EverMind \quad
  \textsuperscript{2}Shanda Group \quad
  \textsuperscript{3}Duke University \\
  \texttt{\{tianwei.lin, zuyi.zhou, xinda.zhao, xiaohong.li,} \\
  \texttt{yu.chen, chuanrui.hu, yafeng.deng\}@shanda.com} \\
  \texttt{cw4565@nyu.edu} \quad \texttt{j.pei@duke.edu}
}

\begin{document}
\maketitle

\input{latex/sec/0_abstract}
\input{latex/sec/1_intro}
\input{latex/sec/2_related}
\input{latex/sec/3_pipeline}

\input{latex/sec/4_exper}
\input{latex/sec/5_limitation}

\bibliography{references}
\clearpage
\input{latex/sec/appendix}

\end{document}

%% file: latex/sec/0_abstract.tex
\begin{abstract}
Long-context LLM agents must access the right evidence from large environments and use it faithfully. However, the popular Needle-in-a-Haystack (NIAH) evaluation mostly measures benign span localization. The needle is near-unique, and the haystack is largely irrelevant. We introduce \textbf{EverMemBench-S (EMB-S)}, an adversarial NIAH-style benchmark built on a \textbf{326M-token MemoryBank}. While the full MemoryBank spans 326M tokens for retrieval-based (RAG) evaluation, we evaluate native long-context models only at scales that fit within each model's context window (up to 1M tokens in this work) to ensure a fair comparison. EMB-S pairs queries with collision-tested near-miss hard negatives and gold evidence sets spanning \emph{one or more} documents, validated via human screening and LLM verification. We also propose a \textbf{decoupled diagnostic protocol} that reports \emph{evidence access} (document-ID localization) separately from end-to-end QA quality under full-context prompting. This enables consistent diagnosis for both native long-context prompting and retrieval pipelines. Across a reference-corpus ladder from domain-isolated 64K contexts to a globally shared 326M-token environment, we observe a clear reality gap. Systems that saturate benign NIAH degrade sharply in evidence access under semantic interference. These results indicate that semantic discrimination, not context length alone, is the dominant bottleneck for long-context memory at scale.
\end{abstract}

%% file: latex/sec/1_intro.tex
\section{Introduction}

% --- Core Figure: EMB-S main contrast ---
\begin{figure}[t]
    \centering
    \includegraphics[width=\columnwidth]{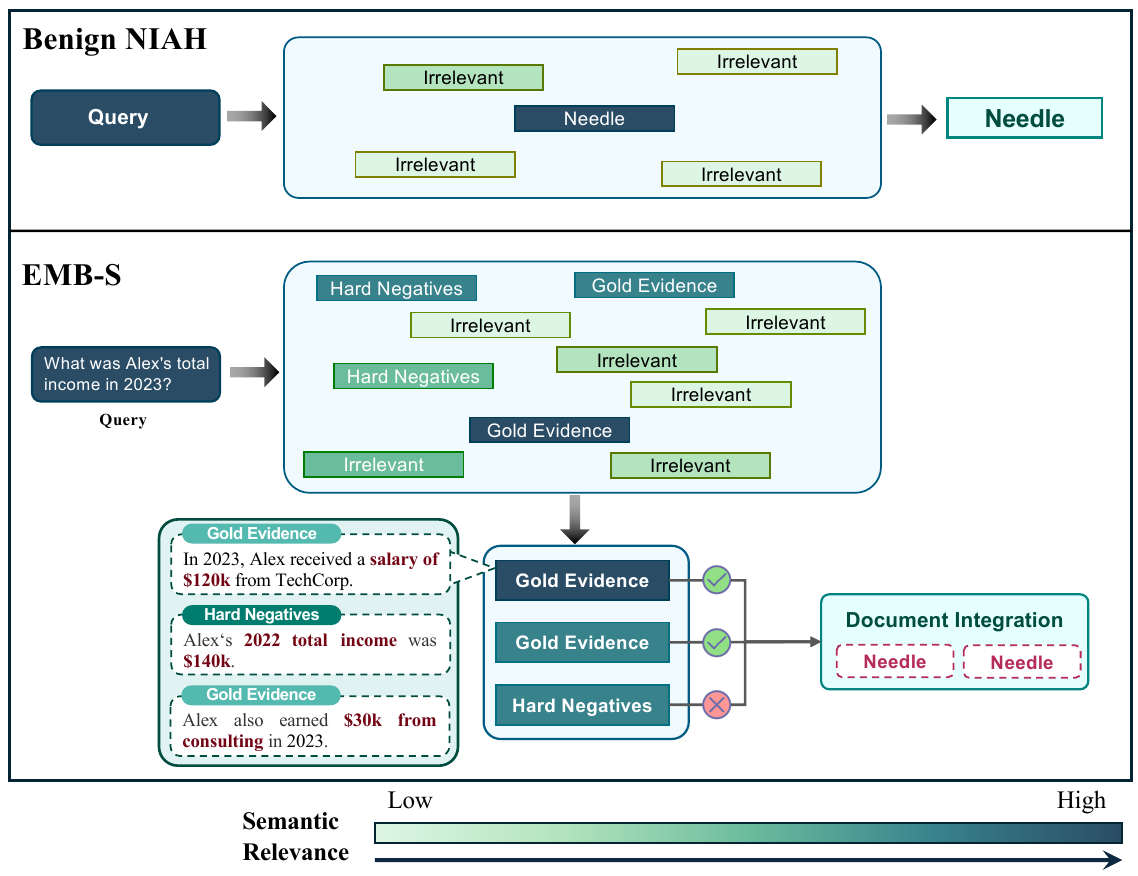}
    \caption{\textbf{Benign NIAH vs.\ EMB-S.} 
    NIAH finds a near-unique needle in mostly irrelevant content; EMB-S must distinguish near-miss negatives and integrate evidence that may span multiple documents.}
    \label{fig:embs_main_core}
\end{figure}

Large language models (LLMs) are increasingly deployed as reasoning layers over large document collections, powering retrieval-augmented generation and tool-using agents in real-world search and analytics systems~\cite{Yehudai2025LLMAgentEval,Ferrag2025LLMAgents}.
In these settings, reliability hinges on evidence-grounded behavior: given a query or internal state, a system must not only access the correct evidence from long contexts or large corpora, but also use it faithfully to produce grounded outputs and actions.

Evaluating this capability at scale, however, remains challenging~\citep{bai2023longbench,an2024leval,bai2024longbenchv2}.
Fully end-to-end evaluation over hundred-million-token environments is computationally expensive and difficult to standardize, which limits reproducibility and broad comparison.
As a result, the community has widely adopted \emph{Needle-in-a-Haystack} (NIAH) tests as a cost-controlled proxy for evidence access~\citep{hsieh2024ruler,kuratov2024babilong,gao2025uniah}.
In a typical NIAH setup, a short ``needle'' is inserted into a long ``haystack,'' and the model is asked to recover it in a single long-context invocation.

However, common NIAH benchmarks primarily measure \emph{benign span localization} rather than realistic evidence access.
In most setups, the needle is (near-)unique and the vast majority of the haystack is irrelevant, so success often reduces to matching a low-entropy string signal~\citep{gao2025uniah}.
Modern long-context models can therefore saturate on NIAH, creating the impression that evidence access has been solved, even though the evaluation rarely stresses semantic ambiguity or competing evidence.

The dominant difficulty in real long-context workloads arises from a different source.
In realistic corpora, evidence is rarely unique: documents overlap, paraphrase one another, and partially satisfy a query while violating a key constraint, such as an incorrect entity, year, or numerical value.
These \emph{near-miss} documents are semantically close to the gold evidence and create dense interference throughout the context.
As a result, evidence access becomes a problem of \textbf{semantic discrimination under global interference}, often requiring the integration of multiple documents rather than the retrieval of a single isolated span.
Crucially, this challenge is orthogonal to context length alone: increasing scale primarily amplifies semantic interference rather than introducing new forms of difficulty.

To bridge this realism gap while preserving the tractability of NIAH-style evaluation, we introduce \textbf{EverMemBench-S (EMB-S)}, an \emph{adversarial NIAH-style} benchmark.
Conceptually, EMB-S is a drop-in upgrade to standard NIAH. It retains single-context evaluation within each model's long-context limits (up to 1M tokens in this work) but replaces largely irrelevant haystacks with collision-tested near-miss hard negatives and gold evidence sets spanning one or more documents, validated via human screening and LLM verification.
As illustrated in Figure~\ref{fig:embs_main_core}, benign NIAH isolates a single needle among irrelevant content. In contrast, EMB-S deliberately mixes multiple gold documents with a dense spectrum of semantically similar distractors. This forces models to discriminate based on constraints rather than simple surface overlap.

Beyond increasing dataset difficulty, EMB-S is paired with a \textbf{decoupled diagnostic protocol} designed to disentangle distinct failure modes.
Specifically, we report \textbf{evidence access} separately from end-to-end QA quality under a shared document-ID evidence interface.
In the \textbf{Localization} task, systems output the top-$k$ document IDs they consider relevant; in \textbf{Generative QA}, we evaluate answer quality under full-context prompting at scales that fit within the model's context window.
This separation helps distinguish access failures from downstream answer-quality degradation under semantic interference, which are otherwise conflated in end-to-end QA metrics~\citep{friel2024ragbench,krishna2024frames,liu2024lostinmiddle}.

We evaluate native long-context models and retrieval-based pipelines across a \textbf{reference corpus ladder}, ranging from \textbf{eight domain-isolated 64K-token corpora} to a \textbf{326M-token MemoryBank}.
This controlled scaling study reveals a consistent pattern: systems that achieve near-perfect scores on benign NIAH degrade sharply in evidence access as semantic interference increases, even when context length is held constant.
These results indicate that semantic discrimination---not raw context length---is the dominant bottleneck for evidence access at scale.

In summary, our contributions are threefold:
\begin{itemize}
    \item \textbf{Adversarial NIAH with semantic interference.} We introduce EverMemBench-S, which augments NIAH with collision-tested near-miss hard negatives and gold evidence sets (single- or multi-document), transforming needle finding into semantic discrimination under dense, constraint-violating distractors.
    \item \textbf{A cost-feasible diagnostic protocol with a shared evidence interface.} EMB-S reports evidence access via document-ID localization metrics and end-to-end QA quality under full-context prompting, enabling diagnosis of access failures and robustness under semantic interference without sacrificing scalability.
    \item \textbf{Unified evaluation across long-context prompting and retrieval pipelines.} EMB-S provides a shared document-ID–based evaluation framework for native long-context prompting, embedding-based retrievers, and retrieve-then-generate systems across corpora ranging from 64K to 326M tokens~\citep{li2024longcontextvsrag,jiang2024longrag}.
\end{itemize}

% Make arXiv/local compilation robust to different working directories.
% - If compiled from within `latex/`, `tab/comparison.tex` is available.
% - If compiled from the project root (e.g., arXiv), `latex/tab/comparison.tex` is available.
\input{latex/tab/comparison}

%% file: latex/tab/comparison.tex
\begin{table*}[t]
\caption{\textbf{Comparison of long-context and memory benchmarks.}
EMB-S uniquely enables decoupled diagnosis of evidence \emph{access} vs.\ \emph{use} in a globally shared environment with dense semantic interference.}
\label{tab:benchmark-comparison}
\centering
\small
\setlength{\tabcolsep}{4pt}
\renewcommand{\arraystretch}{1.2}
%
% KDD/ACM-friendly symbols: avoid heavy color, keep contrast.
\begingroup
\providecommand{\textcolor}[2]{#2} % safe fallback if xcolor isn't loaded
% pifont is loaded in the main tex; we use dingbats for consistent symbols.
\newcommand{\cmark}{\ding{51}}
\newcommand{\xmark}{\ding{55}}
% Use subtle colors for readability (also reasonable in grayscale prints).
\newcommand{\kddyes}{\textcolor{green!60!black}{\cmark}}
\newcommand{\kddno}{\textcolor{red!70!black}{\xmark}}
\begin{tabular*}{\textwidth}{@{\extracolsep{\fill}}@{}lcccccc@{}}
\toprule
\textbf{Dimension}
& \textbf{NIAH}
& \textbf{FRAMES}
& \textbf{BEAM}
& \textbf{LC-Eval}
& \textbf{PRELUDE}
& \textbf{EMB-S} \\
\midrule
\multicolumn{7}{@{}l}{\textbf{\scshape Diagnostic Scope}} \\
\addlinespace[0.1em]
Sparse Evidence Regime        & \kddno & \kddno   & \kddno & \kddno & \kddno & \kddyes \\
Global Shared Env.            & \kddno & \kddyes  & \kddno & \kddno & \kddno & \kddyes \\
Semantic Discrimination       & \kddno & \kddyes  & \kddno & \kddno & \kddno & \kddyes \\
Multi-hop Reasoning           & \kddno & \kddyes  & \kddno & \kddno & \kddno & \kddyes \\
Decoupled Access/Use          & \kddno & \kddno   & \kddno & \kddno & \kddno & \kddyes \\
\addlinespace[0.3em]
\multicolumn{7}{@{}l}{\textbf{\scshape Memory Coverage}} \\
\addlinespace[0.1em]
Working Memory (Context)      & \kddyes & \kddyes & \kddyes & \kddyes & \kddyes & \kddyes \\
Parametric Memory             & \kddyes & \kddno  & \kddyes & \kddyes & \kddyes & \kddyes \\
External Memory               & \kddno  & \kddyes & \kddyes & \kddno  & \kddno  & \kddyes \\
Persistent/Latent             & \kddno  & \kddno  & \kddno  & \kddno  & \kddno  & \kddyes \\
\addlinespace[0.3em]
\multicolumn{7}{@{}l}{\textbf{\scshape Scale and Realism}} \\
\addlinespace[0.1em]
Verified Hard Negatives       & \kddno  & \kddyes & \kddno  & \kddno  & \kddno  & \kddyes \\
Scale $>1$M Tokens            & \kddyes & \kddno  & \kddyes & \kddno  & \kddno  & \kddyes \\
Multi-scale Eval.             & \kddyes & \kddno  & \kddyes & \kddyes & \kddno  & \kddyes \\
Diverse Corpus                & \kddno  & \kddyes & \kddyes & \kddyes & \kddno  & \kddyes \\
\bottomrule
\end{tabular*}
\endgroup
\end{table*}

%% file: latex/sec/2_related.tex
% ============================================
% 2. Related Work
% ============================================
\section{Related Work}

\paragraph{Evidence Access under Global Interference.}
Long-context evaluation suites such as LongBench~\citep{bai2023longbench}, L-Eval~\citep{an2024leval}, LV-Eval~\citep{yuan2024lveval}, LongBench v2~\citep{bai2024longbenchv2}, RULER~\citep{hsieh2024ruler}, and BABILong~\citep{kuratov2024babilong} have driven progress on end-to-end QA and reasoning over extended contexts.
These benchmarks primarily evaluate end-to-end task success, implicitly assuming sparse relevance and near-unique evidence, and therefore do not stress semantic discrimination under globally shared, hard-negative-heavy contexts.
A key limitation is that end-to-end accuracy conflates (i) failing to access the evidence, (ii) accessing it but being diluted by interference, and (iii) accessing the right evidence but making a reasoning/grounding error.
A key limitation in benchmark construction is that relevance is typically sparse and evidence near-unique, without explicitly constructing constraint-violating near-miss distractors, so these suites largely measure \emph{length}, not \emph{difficulty}.

\paragraph{From Needle Localization to Semantic Discrimination.}
NIAH-style needle insertion tests offer a cheap and scalable probe of long-context behavior by planting a short target span in a long haystack and asking models to recover it~\citep{hsieh2024ruler,kuratov2024babilong,gao2025uniah}.
A key limitation is that the needle is often a low-entropy, near-unique signal, with few near-miss candidates, no systematic constraint-violation design, and an implicit single-needle assumption.
Under dense near-miss conditions, evidence access ceases to be a localization problem and becomes semantic discrimination under global interference.

\paragraph{Decoupling Access and Use at Scale.}
RAG benchmarks such as RAGBench~\citep{friel2024ragbench}, FRAMES~\citep{krishna2024frames}, CRUD-RAG~\citep{lyu2024crudrag}, and DomainRAG~\citep{wang2024domainrag} decouple retrieval and generation and provide component-level metrics for pipeline-based systems.
Recent work also revisits the long-context vs.\ RAG trade-off and explores long-context-enhanced retrieval pipelines~\citep{li2024longcontextvsrag,jiang2024longrag}.
These benchmarks decouple retrieval and generation, but are architecturally tied to pipeline-based systems, preventing unified evaluation of native long-context prompting and retrieval-based systems under the same evidence interface.
A key limitation is the lack of a shared document-ID evidence space that supports architecture-agnostic diagnosis of evidence access under large, globally mixed corpora.

\paragraph{Positioning \& Summary.}
A key gap is scalable evaluation of evidence access under dense semantic interference, where answers require multi-evidence aggregation and near-miss distractors are both abundant and adversarial.
EMB-S occupies a previously underexplored regime: scalable evaluation of evidence access under dense semantic interference, with explicit diagnostics that separate access from use.

Table~\ref{tab:benchmark-comparison} summarizes how EMB-S differs along these diagnostic dimensions.

%% file: latex/sec/3_pipeline.tex
\section{Dataset Construction}
\label{sec:pipeline}

\begin{figure*}[!t]
    \centering
    \includegraphics[width=\textwidth]{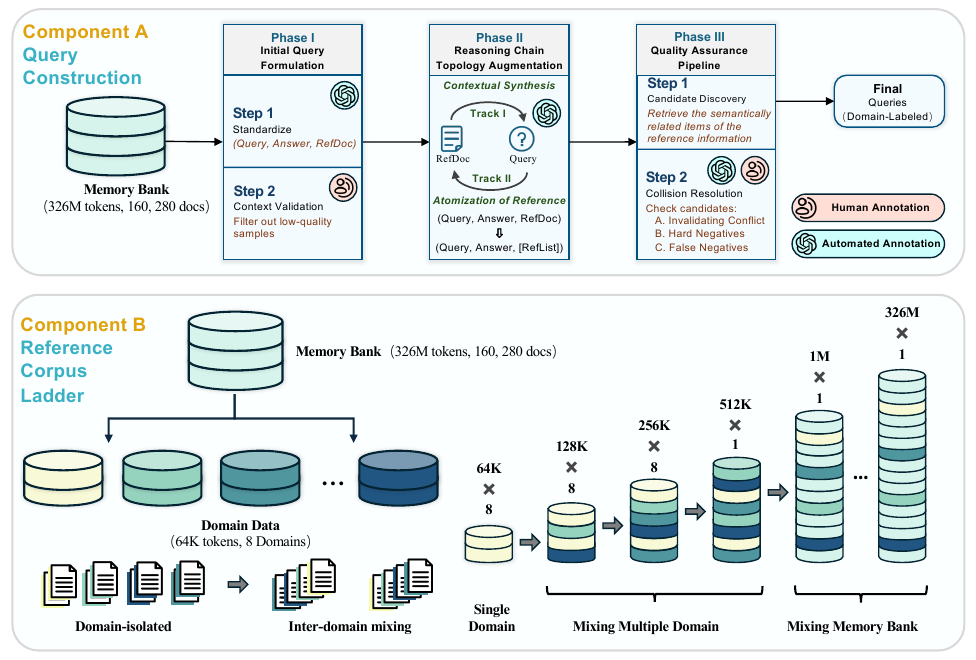}
    \caption{\textbf{EMB-S overview.} Component A builds 483 domain-labeled queries from a 326M-token MemoryBank (standardization, multi-hop synthesis, and collision testing). Component B defines a reference-corpus ladder from 64K to 326M tokens with increasing inter-domain mixing and distractor injection.}
    \label{fig:overall_structure}
\end{figure*}

Figure~\ref{fig:overall_structure} provides an overview of EMB-S's data foundation and construction pipeline. Our key design choice is to \emph{fix the query set} while \emph{systematically varying the searchable evidence pool}: Component A constructs a compact set of rigorously validated, domain-labeled queries with gold evidence sets (\textbf{RefList}) spanning one or more documents and collision-tested distractors (via human screening and LLM verification), while Component B defines a multi-scale reference-corpus ladder that increases both \emph{scale} and \emph{semantic interference} in a controlled manner. This decoupling enables apples-to-apples comparisons across native long-context prompting and retrieval-based pipelines under matched supervision.

\subsection{Raw Data Sources}
\label{sec:data-sources}

The construction of EverMemBench-S begins with systematically collecting existing, publicly available long-context evaluation benchmarks. We aggregate 9 diverse datasets spanning different scales, task types, and evaluation objectives to form the 326M-token MemoryBank (160,280 documents); the complete list of sources and their characteristics is provided in \textbf{Appendix~\ref{sec:appendix-sources}}. All raw datasets are downloaded, cleaned, and deduplicated before further processing \citep{magar2022data}. To enable domain-conditioned evaluation at long-context scale, we additionally assign documents (and downstream queries) to one of 8 broad domains; details and guidelines are provided in Appendix~\ref{sec:appendix-structure}.

\subsection{Data Standardization}
\label{sec:data-standardization}

To ensure cross-source consistency, we transform all raw instances into a unified tuple: \textbf{(Query, Answer, RefDoc)}. Here, \textbf{RefDoc} denotes the reference document that must be accessed to correctly answer the query. In later stages, we form a gold evidence set \textbf{RefList} spanning one or more documents; we call a query \textbf{single-source} if $|\text{RefList}|=1$ (i.e., it requires retrieving a single gold reference document) and \textbf{multi-source} otherwise. We also collect a pool of hard negatives for adversarial evaluation under semantic interference.

\subsection{Data Construction Overview}

EverMemBench-S is constructed via a systematic process with two tightly coupled components (Figure~\ref{fig:overall_structure}). \textbf{Component A} applies a three-stage query construction pipeline that transforms 39,860 heterogeneous instances into \textbf{483} rigorously validated, domain-labeled queries with gold evidence sets (\textbf{RefList}) and collision-tested distractors (Table~\ref{tab:pipeline-stats}). \textbf{Component B} then builds a multi-scale \emph{reference corpus ladder} that starts from domain-isolated long-context corpora and expands to the full 326M-token MemoryBank via inter-domain mixing and progressive distractor injection (1M/2M/\dots/326M). We provide additional implementation details (e.g., stage-level criteria, domain labeling, and verification guidelines) in Appendix~\ref{sec:appendix-structure}.

\input{latex/tab/quantitative}

\textbf{Stage I: Standardization and Human Screening.}
\label{sec:purification}
Stage~I converts heterogeneous sources into standardized triples \textbf{(Query, Answer, RefDoc)}, where RefDoc is the reference document required to infer the answer from the query. We then perform \emph{human screening} to remove low-quality instances (e.g., ambiguous questions, unsupported answers, or noisy/mismatched references), yielding a clean seed pool for downstream multi-hop synthesis.

\textbf{Stage II: Reasoning-Chain Synthesis (Two Tracks).}
\label{sec:augmentation}
Stage~II converts single-hop triples into reasoning-intensive items and constructs the gold evidence set \textbf{RefList}. \textbf{RefList} may remain single-document (single-source) or be expanded into multiple documents (multi-source). To create multi-source items, we implement two complementary expansion tracks: \textit{Track 2 (RefDoc atomization)} decomposes complex reference documents into multiple sub-documents, forcing cross-document aggregation; \textit{Track 1 (retrieval-guided query rewriting)} retrieves semantically similar documents from the MemoryBank and rewrites the query so that answering requires both the original RefDoc and newly added supporting documents. We then calibrate difficulty and remove near-duplicates using a strong dense retriever (\textbf{Qwen3-Embedding-8B}), producing 882 candidate queries for Stage~III (Table~\ref{tab:pipeline-stats}).

\textbf{Stage III: Quality Control via Collision Testing.}
\label{sec:collision-testing}
Stage~III improves scoring reliability by filtering inconsistencies and validating negatives. For each candidate triple, we retrieve the top-$k$ nearest \emph{non-reference} documents under dense embedding similarity from the MemoryBank (collision candidates), using \textbf{Qwen3-Embedding-8B} (excluding documents in \textbf{RefList}). An LLM then verifies each candidate against \textbf{(Query, Answer, RefList)} and assigns one of three outcomes: \textbf{(a) Conflict} (contradicts the query/answer/reference content; discard the sample), \textbf{(b) Hard Negative} (semantically similar but does not support the answer; retained as an adversarial distractor), or \textbf{(c) False Negative} (provides additional valid evidence; added to RefList). The final benchmark contains \textbf{483} queries with LLM-verified reference sets and validated hard negatives.

\subsection{Reference Corpus Ladder}
\label{sec:corpus-ladder}

Beyond constructing queries, EMB-S defines a \emph{reference corpus ladder} that enables controlled stress tests under increasing \emph{scale} and \emph{cross-domain interference} while keeping the query set fixed (Figure~\ref{fig:overall_structure}). Each query is assigned a domain label $d \in \{1,\ldots,8\}$ based on the domains of its gold reference documents, and we construct a family of reference corpora $\{\mathcal{C}\}$ as follows.
\textbf{Scale as token budget.} We parameterize each corpus by a target token budget $S$ (e.g., 64K, 512K, 326M), i.e., the total tokens of the searchable evidence pool $\mathcal{C}^S$ rather than the number of documents.\footnote{When serializing $\mathcal{C}^S$ into a single input for a model with window $N$, we additionally enforce the constraint using the model's tokenizer (Appendix~\ref{sec:eval-scenarios}).}

\textbf{Domain-isolated base corpora (64K).} For each domain $d$, we build a domain-specific corpus $\mathcal{C}_d^{64K}$ that (i) contains all gold reference documents for queries in domain $d$ and (ii) fits within a 64K long-context input after accounting for prompt/query overhead. This setting supports \emph{domain-isolated} evaluation, where each domain can be tested independently.

\textbf{Inter-domain mixing (128K/256K).} To gradually introduce semantic interference while retaining a domain-conditioned evaluation axis, we expand each domain corpus by sampling documents from the other 7 domains. Let $\mathcal{S}(\mathcal{D}, B)$ denote sampling documents from a collection $\mathcal{D}$ until the total token budget reaches $B$:
\begin{align*}
    \mathcal{C}_d^{S} &= \mathcal{C}_d^{64K} \cup \mathcal{S}\left(\bigcup_{d' \neq d}\mathcal{C}_{d'}^{64K};\, S-64\mathrm{K}\right) \\
    &\text{for } S \in \{128\mathrm{K}, 256\mathrm{K}\},
\end{align*}
where sampling is performed without replacement and with document-level deduplication.

\textbf{Shared mid-scale corpus (512K).} We then merge the 8 domain pools into a shared corpus $\mathcal{C}^{512K}=\mathrm{Dedup}\!\left(\bigcup_{d=1}^{8}\mathcal{C}_d^{64K}\right)$, yielding a single reference space shared by all domains.

\textbf{Global distractor injection.} Finally, we expand the shared corpus by injecting distractor documents sampled from the remaining MemoryBank (denoted as $\mathcal{M}$) to reach larger scales (e.g., 1M, 2M, \dots, up to 326M tokens):
\begin{align*}
    \mathcal{C}^{S} = \mathcal{C}^{512K} \cup \mathcal{S}\big(\mathcal{M} \setminus \mathcal{C}^{512K};\, S-512\mathrm{K}\big),
\end{align*}
with a fixed random seed to ensure reproducibility across runs and scales.

% Table summarizing the entire pipeline statistics
\subsection{Dataset Characteristics and Sparsity Validation}
\label{sec:dataset-characteristics}

\begin{figure}[!htbp]
    \centering
    \includegraphics[width=0.95\columnwidth]{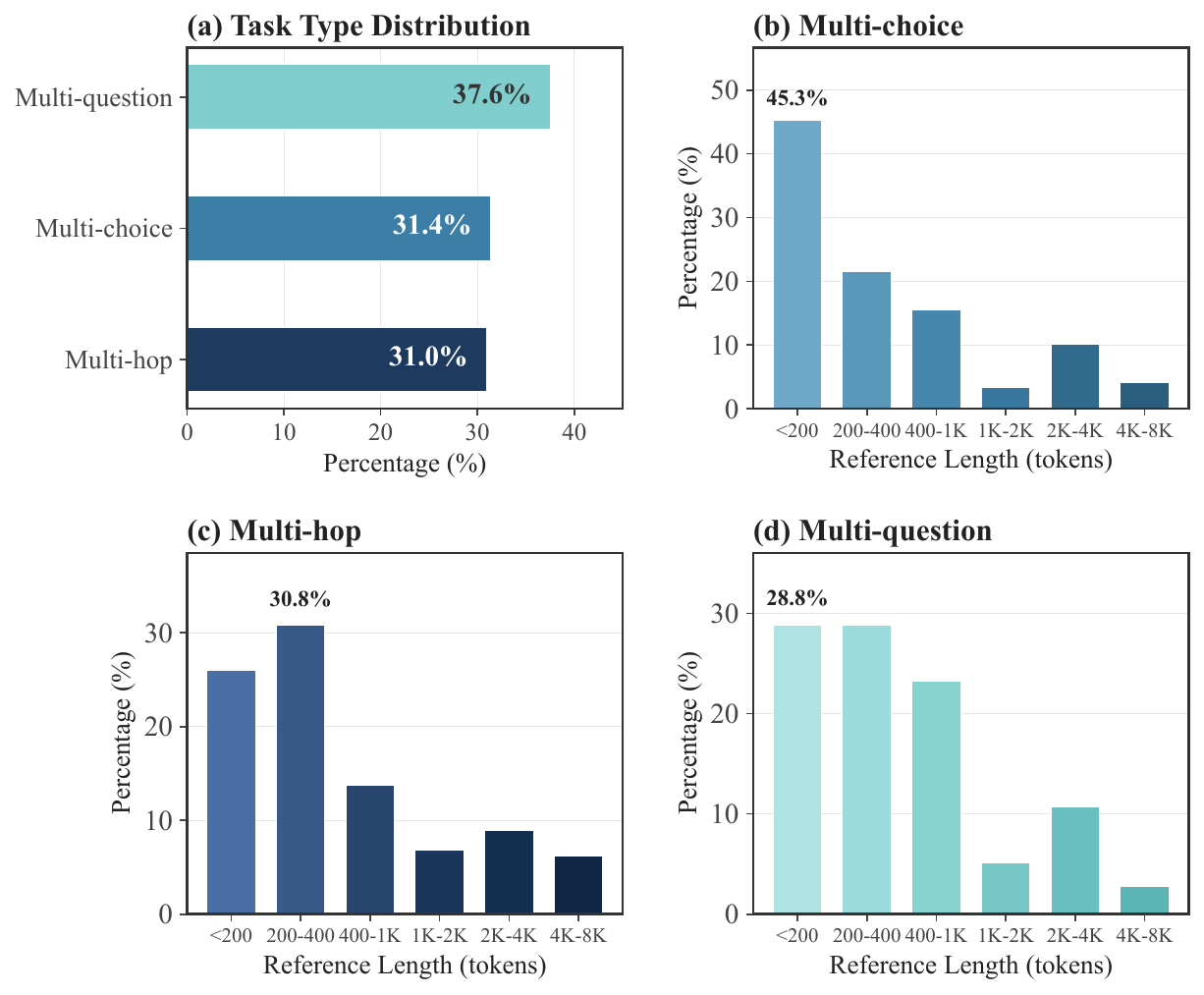}
    \caption{Distribution of task types in EverMemBench-S and the distribution of context lengths for each task type.}
    \label{fig:task_distribution}
\end{figure}

\begin{figure}[!htbp]
    \centering
    \includegraphics[width=0.95\columnwidth]{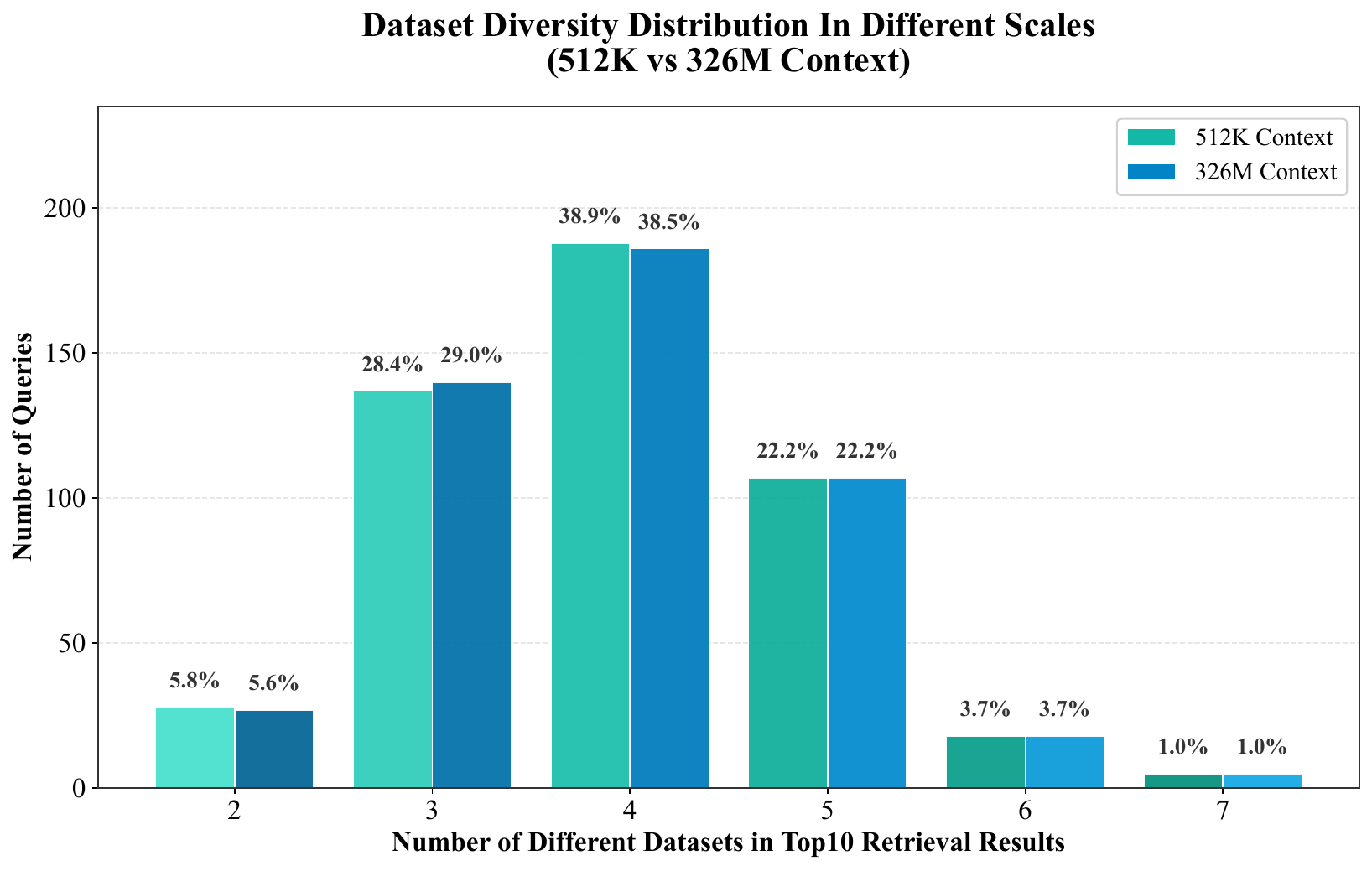}
    \caption{\textbf{Cross-source mixing (512K vs.\ 326M).} Distribution of the number of distinct datasets in the top-10 retrieved documents per query.}
    \label{fig:dataset_diversity}
\end{figure}

We summarize two key properties of the constructed benchmark that directly impact evaluation difficulty: (i) the distribution of query types and context lengths (Figure~\ref{fig:task_distribution}), and (ii) whether the reference corpus ladder induces domain ``silos'' or a globally mixed search space under scaling (Figure~\ref{fig:dataset_diversity}).

\textbf{Validation of Interference Progression.} Figure~\ref{fig:dataset_diversity} shows that retrieval becomes more cross-source as the corpus scales (e.g., 512K$\rightarrow$326M): top-$k$ results increasingly span multiple datasets, indicating a shift toward a more globally mixed retrieval environment.

%% file: latex/tab/quantitative.tex
\begin{table}[t]
    \caption{\textbf{Pipeline statistics.} Quantitative overview of the EverMemBench-S construction pipeline (instance counts across stages; final benchmark contains \textbf{483} queries).}
    \label{tab:pipeline-stats}
    \centering
    \small
    \setlength{\tabcolsep}{3.5pt}
    \renewcommand{\arraystretch}{1.12}
    \begin{tabular}{@{}llr@{}}
    \toprule
    \textbf{Stage} & \textbf{Step / Process} & \textbf{\#Inst.} \\
    \midrule
    \multicolumn{3}{@{}l}{\textbf{\scshape Stage I: Standardization \& human screening}} \\
    \addlinespace[0.15em]
    & \hspace{0.6em}Raw aggregation & 39,860 \\
    & \hspace{0.6em}Human screening (seed pool) & 9,621 \\
    \addlinespace[0.35em]
    \multicolumn{3}{@{}l}{\textbf{\scshape Stage II: Reasoning-chain synthesis}} \\
    \addlinespace[0.15em]
    & \hspace{0.6em}Seed pool (from Stage I) & 9,621 \\
    & \hspace{0.6em}Track 1: Retrieval-guided query rewriting & 2,112 \\
    & \hspace{0.6em}Track 2: RefDoc atomization & 7,509 \\
    & \hspace{0.6em}Synthesis output (merged) & 3,457 \\
    & \hspace{0.6em}Deduplication \& difficulty calibration & 882 \\
    \addlinespace[0.35em]
    \multicolumn{3}{@{}l}{\textbf{\scshape Stage III: Quality control (collision testing)}} \\
    \addlinespace[0.15em]
    & \hspace{0.6em}Collision testing \& LLM verification (final) & \textbf{483} \\
    \bottomrule
    \end{tabular}
    \end{table}

%% file: latex/sec/4_exper.tex
% ============================================
% 5. Experimental Results and Analysis
% ============================================
\section{Experiments}
\label{sec:experiments}

\input{latex/tab/result}

\subsection{Research Questions}
\label{sec:exper-rq}
We structure our experiments as a diagnostic study centered on \emph{evidence access under interference} and its downstream impact on long-context question answering. Specifically, we investigate the following research questions:

\begin{itemize}[leftmargin=*, itemsep=1pt, parsep=0pt]
    \item RQ1 (Scaling): How does \emph{evidence access} degrade as the searchable corpus scales and semantic interference increases?
    \item RQ2 (Single vs.\ multi-source): How large is the gap between retrieving a single required reference document and retrieving \emph{all} required reference documents under interference?
    \item RQ3 (End-to-end QA): Under realistic long-context usage, how well can long-context LLMs answer questions when provided the full reference corpus within their context-window constraints?
\end{itemize}

\subsection{Experimental Setup}
\label{sec:exper-setup}
We evaluate 483 queries over a reference corpus ladder that scales from 64K (domain-isolated) through 128K/256K and a shared 512K corpus, up to a 326M-token MemoryBank with global distractor injection. Throughout, we distinguish the \emph{environment scale} $S$ (the size of the searchable evidence pool $\mathcal{C}^S$, up to 326M tokens) from the \emph{model input scale} $N$ (a model's maximum context length per invocation). For evidence access, we retrieve top-$K$ document IDs from $\mathcal{C}^S$ and report R@1 (single-source), SR@10, and FR@10 (strict multi-source). For end-to-end QA, we answer queries given the full reference corpus as context whenever the full input fits within $N$, scored by LLM-as-a-Judge on a 0--5 scale (Grok-4; prompt in Figure~\ref{fig:judge-prompt}).
We adopt a \emph{diagnostic} protocol with a shared document-ID evidence interface that reports evidence access separately from end-to-end QA quality. Evidence access is evaluated via document-ID localization/recall over $\mathcal{C}^S$, while QA scores are reported under full-context prompting when feasible. Appendix~\ref{sec:docid-protocol} defines the document-ID interface and LLM output canonicalization.

We define a query as single-source if its gold evidence set contains exactly one reference document, and multi-source otherwise.

Evaluation regimes.
(i) For RAG-style systems, we evaluate \emph{evidence access} only, measuring whether the retriever surfaces the gold reference documents as the searchable corpus scales along the ladder.
(ii) For native long-context LLMs, we evaluate end-to-end QA by scoring answers produced with the full reference corpus provided as context whenever it fits within the model's context window $N$.
Appendix~\ref{sec:eval-details} provides full implementation details.

\subsection{Evidence Access under Semantic Interference (RQ1--RQ2)}
\label{sec:retrieval-results}
This section evaluates how well retrievers localize gold evidence as corpus scale and interference increase. We report results for dense retrievers, a sparse BM25 baseline, and a reranking-enhanced variant as an auxiliary analysis.

\begin{figure}[t]
    \centering
    \includegraphics[width=\columnwidth]{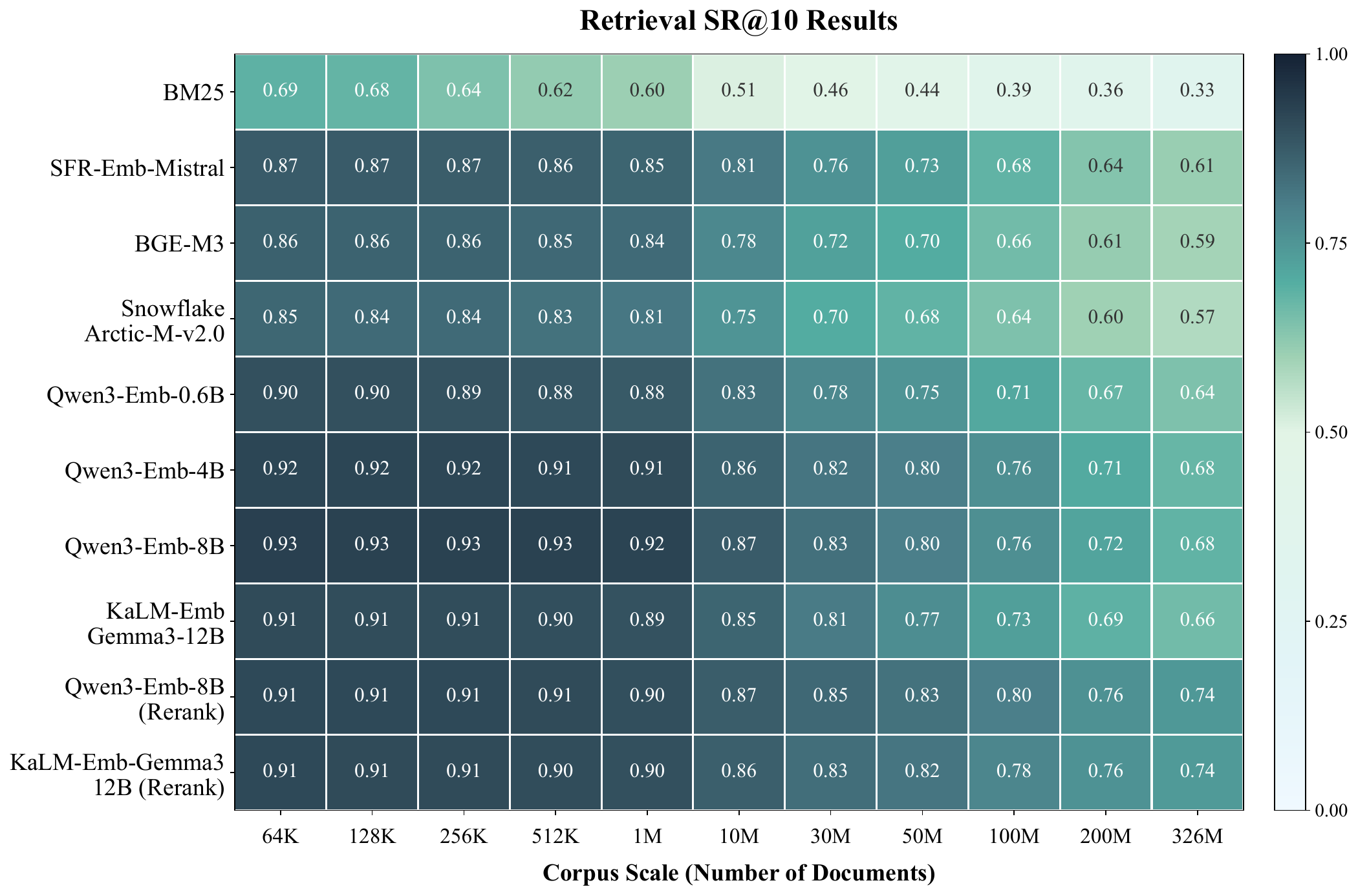}
\caption{Retrieval heatmap across the reference corpus ladder (64K--326M): evidence access degrades as the searchable pool expands and semantic interference increases.}
    \label{fig:heatmap}
\end{figure}

Figure~\ref{fig:heatmap} visualizes SR@10 degradation across the full corpus ladder, while Table~\ref{tab:retrieval_results} reports representative scales (64K/512K/30M/326M) with R@1, SR@10, and FR@10.

As corpus scale increases from 64K to 326M, evidence access degrades substantially across all retrievers. For example, a strong dense retriever drops from SR@10 $\approx 0.93$ at 64K to $\approx 0.68$ at 326M, while FR@10 declines more sharply (e.g., $\approx 0.80 \rightarrow 0.30$). In contrast, BM25 degrades rapidly under interference, with FR@10 approaching zero at large scale.

Across embedding models, Qwen3-Embedding-8B attains the best SR@10 and FR@10 at all four reported scales (e.g., SR@10 0.930$\rightarrow$0.682 and FR@10 0.800$\rightarrow$0.304 from 64K to 326M), while KaLM-Embedding-Gemma3-12B yields the strongest R@1 at 64K/512K (0.905/0.892) and ties Qwen3-Embedding-8B at 326M (0.622). The SR@10--FR@10 gap widens from 0.13 at 64K to 0.38 at 326M for Qwen3-Embedding-8B, underscoring multi-source brittleness.

\paragraph{Single-source vs.\ multi-source retrieval (RQ2).}
Across all scales, FR@10 is consistently far below SR@10, and the gap widens as interference increases. This demonstrates that retrieving \emph{all} required evidence documents is substantially harder than retrieving a single relevant document, making multi-source queries a strict stress test for evidence access.

\paragraph{Auxiliary analysis: reranking as mitigation.}
While not a primary benchmark axis, reranking provides a partial mitigation under heavy interference. At 326M, reranking lifts Qwen3-Embedding-8B from SR@10 0.682 to 0.745 and FR@10 0.304 to 0.400; for KaLM-Embedding-Gemma3-12B, SR@10 improves from 0.657 to 0.739 and FR@10 from 0.263 to 0.391. This suggests that hard-negative discrimination becomes increasingly valuable as corpus scale grows.

\subsection{End-to-end Long-context QA under Full Context (RQ3)}
\label{sec:qa-results}
We evaluate end-to-end question answering using LLM-as-a-Judge scores (0--5) under full-context prompting. We report results on representative scales (64K/256K/1M; Table~\ref{tab:qa_results}) only when the full input fits within each model's context window (up to 1M here); missing entries indicate that the full input at that scale exceeds the model's context window and thus cannot be evaluated. The best scores come from Gemini-3-Pro-Preview at 64K/256K/1M (3.55/3.45/3.28).

Two consistent patterns emerge. First, answer quality can drop as the full-context scale increases (e.g., Gemini-3-Pro-Preview 3.55$\rightarrow$3.28 from 64K to 1M), reflecting the increasing difficulty of full-context QA under heavier semantic interference. Second, larger context windows alone do not guarantee better answer quality: models with comparable or larger windows can still trail behind the best-performing model at the same evaluated scales.

\subsection{Key Takeaways}
\label{sec:insights}
Our experiments yield three takeaways aligned with RQ1--RQ3.
(1) Scaling induces access failures: evidence access degrades substantially as corpus scale and interference increase.
(2) Multi-source retrieval is the critical stress point: the widening SR@10--FR@10 gap shows that satisfying multiple evidence constraints is particularly fragile.
(3) Long-context capacity is not sufficient: even when full-context evaluation is feasible, full-context QA quality can degrade with scale and varies substantially across models.
Together, these findings demonstrate that success on benign long-context settings does not predict robustness in high-interference environments, and that effective long-context systems require robust evidence access paired with strong reasoning.

%% file: latex/tab/result.tex
% -----------------------------
% Table A: Retrieval results
% -----------------------------
\begin{table*}[!htbp]
    \caption{\textbf{Retrieval performance (RAG / evidence access).}
    Results are reported on four representative scales on the reference-corpus ladder: 64K, 512K, 30M, and 326M.
    We report \textbf{R@1} (single-source queries only), \textbf{SR@10} (standard recall@10 over all queries), and \textbf{FR@10} (strict recall@10 on multi-source queries requiring all references). We additionally report reranking results (Qwen3-Reranker-8B) where available.}
    \label{tab:retrieval_results}
    \centering
    \scriptsize
    \setlength{\tabcolsep}{2.6pt}
    \renewcommand{\arraystretch}{1.08}
    \providecommand{\rowindent}{\hspace{0.6em}}
    \begin{tabular*}{\textwidth}{@{\extracolsep{\fill}}@{}lcccccccccccc@{}}
    \toprule
    \textbf{Model} &
    \multicolumn{3}{c}{\textbf{64K}} &
    \multicolumn{3}{c}{\textbf{512K}} &
    \multicolumn{3}{c}{\textbf{30M}} &
    \multicolumn{3}{c}{\textbf{326M}} \\
    \cmidrule(lr){2-4}\cmidrule(lr){5-7}\cmidrule(lr){8-10}\cmidrule(lr){11-13}
    & \textbf{R@1} & \textbf{SR@10} & \textbf{FR@10}
    & \textbf{R@1} & \textbf{SR@10} & \textbf{FR@10}
    & \textbf{R@1} & \textbf{SR@10} & \textbf{FR@10}
    & \textbf{R@1} & \textbf{SR@10} & \textbf{FR@10} \\
    \midrule
    \multicolumn{13}{@{}l}{\textbf{\textit{Embedding models}}} \\
    \addlinespace[0.25em]
    \rowindent SFR-Embedding-Mistral & 0.858 & 0.874 & 0.648 & 0.845 & 0.864 & 0.621 & 0.750 & 0.755 & 0.436 & 0.554 & 0.607 & 0.209 \\
    \rowindent BGE-M3 & 0.831 & 0.861 & 0.642 & 0.811 & 0.847 & 0.615 & 0.696 & 0.724 & 0.397 & 0.480 & 0.591 & 0.173 \\
    \rowindent Snowflake-Arctic-M-v2.0 & 0.831 & 0.848 & 0.609 & 0.818 & 0.827 & 0.573 & 0.736 & 0.703 & 0.361 & 0.446 & 0.572 & 0.173 \\
    \rowindent Qwen3-Embedding-8B & 0.872 & \textbf{0.930} & \textbf{0.800} & 0.865 & \textbf{0.928} & \textbf{0.788} & 0.804 & 0.831 & 0.585 & 0.622 & 0.682 & 0.304 \\
    \rowindent Qwen3-Embedding-4B & 0.872 & 0.921 & 0.767 & 0.858 & 0.912 & 0.749 & 0.804 & 0.818 & 0.561 & 0.588 & 0.675 & 0.304 \\
    \rowindent Qwen3-Embedding-0.6B & 0.851 & 0.897 & 0.716 & 0.831 & 0.884 & 0.684 & 0.784 & 0.778 & 0.478 & 0.568 & 0.644 & 0.248 \\
    \rowindent KaLM-Embedding-Gemma3-12B & \textbf{0.905} & 0.908 & 0.734 & \textbf{0.892} & 0.899 & 0.713 & 0.818 & 0.806 & 0.540 & 0.622 & 0.657 & 0.263 \\
    \addlinespace[0.35em]
    \multicolumn{13}{@{}l}{\textbf{\textit{Sparse retrieval}}} \\
    \addlinespace[0.25em]
    \rowindent BM25 & 0.655 & 0.687 & 0.358 & 0.622 & 0.616 & 0.266 & 0.514 & 0.461 & 0.116 & 0.345 & 0.327 & 0.030 \\
    \addlinespace[0.35em]
    \multicolumn{13}{@{}l}{\textbf{\textit{Rerank (Qwen3-Reranker-8B)}}} \\
    \addlinespace[0.25em]
    \rowindent Qwen3-Embedding-8B & 0.872 & 0.913 & 0.770 & 0.865 & 0.908 & 0.758 & 0.818 & \textbf{0.846} & \textbf{0.600} & \textbf{0.689} & \textbf{0.745} & \textbf{0.400} \\
    \rowindent KaLM-Embedding-Gemma3-12B & 0.885 & 0.911 & 0.761 & 0.865 & 0.903 & 0.740 & \textbf{0.838} & 0.834 & 0.567 & 0.676 & 0.739 & 0.391 \\
    \bottomrule
    \end{tabular*}
\end{table*}
    
    % -----------------------------
    % Table B: QA (LLM-as-a-Judge)
    % -----------------------------
\providecommand{\rowindent}{\hspace{0.6em}}
\begin{table}[t]
    \caption{\textbf{Generative QA quality (LLM-as-a-Judge, 0--5).}
    Full-context results at 64K/256K/1M when feasible.}
    \label{tab:qa_results}
    \centering
    \scriptsize
    \setlength{\tabcolsep}{2.1pt}
    \renewcommand{\arraystretch}{1.1}
    \begin{tabular*}{\columnwidth}{@{\extracolsep{\fill}}@{}p{0.46\columnwidth}cccc@{}}
    \toprule
    \textbf{Model} & \textbf{Context window} & \textbf{64K} & \textbf{256K} & \textbf{1M} \\
    \midrule

    \multicolumn{5}{@{}l}{\textbf{\textit{Full-capacity Models}}} \\
    \addlinespace[0.25em]
    \rowindent Qwen3-235B-A22B-Instruct & 256K & 3.35 & 3.20 & - \\
    \rowindent Qwen3-Next-80B & 256K & 3.35 & 3.30 & - \\
    \rowindent DeepSeek-V3.2 & 128K & 3.40 & - & - \\
    \rowindent GLM-4.7 & 200K & 3.18 & - & - \\
    \rowindent Gemini-3-Pro-Preview & 1M & \textbf{3.55} & \textbf{3.45} & \textbf{3.28} \\

    \addlinespace[0.35em]
    \multicolumn{5}{@{}l}{\textbf{\textit{Efficiency-oriented Models}}} \\
    \addlinespace[0.25em]
    \rowindent GLM-4.7-Flash & 200K & 3.10 & - & - \\
    \rowindent MiniMax-M2.1 & 204K & 3.38 & - & - \\
    \rowindent Grok-4.1-Fast & 2M & 3.45 & 3.36 & 3.01 \\
    \rowindent Kimi-Linear-48B-A3B-Instruct & 1M & 3.42 & 3.38 & 3.22 \\
    \rowindent Gemini-3-Flash-Preview & 1M & \textbf{3.48} & \textbf{3.40} & \textbf{3.25} \\

    \bottomrule
    \end{tabular*}
\end{table}

%% file: latex/sec/5_limitation.tex
% ============================================
% 5. Limitations
% ============================================
\section*{Limitations}
\label{sec:limitations}

While this work establishes a rigorous benchmark for large-scale memory, four limitations warrant further discussion. \textbf{First}, EMB-S prioritizes high-precision instance validation and hard-negative verification, which yields a relatively compact set; with additional expert time and expanded source pools, the same pipeline can scale to broader coverage and more diverse failure modes. \textbf{Second}, our document-ID localization protocol for native LLMs (Appendix~\ref{sec:docid-protocol}) introduces an explicit evidence interface that may not reflect typical long-context usage; despite output canonicalization, this setup can introduce formatting and calibration biases when comparing against dedicated retrievers. \textbf{Third}, parts of our construction and evaluation rely on specific tools and models (e.g., Grok-4 as the judge; Qwen3-Embedding-8B for difficulty calibration and collision testing); while we mitigate this with human screening and human evaluation, future work should replicate results with alternative judges and multi-retriever mining. \textbf{Fourth}, we do not account for inference-time efficiency gaps between full-context prompting and retrieval pipelines (e.g., latency and GPU memory), which can be substantial at million-token inputs; EMB-S is intended to diagnose robustness under semantic interference rather than to advocate one paradigm over the other.

%% file: latex/sec/appendix.tex
\appendix
% Appendix text should remain two-column under ACL style.
\section{Implementation Details}
\label{sec:eval-details}

\subsection{Construction Pipeline Details}
\label{sec:appendix-structure}

\noindent\textbf{Overview.} Figure~\ref{fig:overall_structure} (main paper) summarizes the end-to-end construction pipeline, and Table~\ref{tab:pipeline-stats} reports the instance counts across stages. This appendix provides additional details omitted from the main text, including stage-level criteria, human screening guidelines, and LLM verification criteria.

\subsubsection{Stage I: Standardization and Human Screening}
\noindent Stage~I converts heterogeneous sources into standardized triples and removes low-quality items before multi-hop synthesis. We implement two steps:
\begin{itemize}[leftmargin=*, itemsep=1pt, parsep=0pt]
    \item \textit{Step 1 (Standardization):} convert raw instances into \textbf{(Query, Answer, RefDoc)}, where RefDoc is the reference document that must be accessed to derive the answer from the query.
    \item \textit{Step 2 (Human screening):} validate that the query is well-posed and that RefDoc supports the answer; discard ambiguous, unsupported, or noisy instances.
\end{itemize}

\subsubsection{Stage II: Reasoning-Chain Synthesis (Two Tracks)}
\noindent Stage~II constructs reasoning-intensive stress tests by constructing the gold evidence set \textbf{RefList}, which may contain one document (single-source) or multiple documents (multi-source). To create multi-source items, we expand RefDoc into a multi-document RefList via two complementary tracks:
\begin{itemize}[leftmargin=*, itemsep=1pt, parsep=0pt]
    \item \textit{Track 2 (RefDoc atomization):} for RefDoc with complex internal logic, we \emph{atomize} it into multiple sub-documents so that no single piece is sufficient.
    \item \textit{Track 1 (Retrieval-guided query rewriting):} retrieve semantically similar documents from the MemoryBank, select related documents that provide complementary constraints, and rewrite the query so that answering requires both the original RefDoc and newly added supporting documents.
\end{itemize}
\noindent We then perform difficulty calibration and deduplication using a strong dense retriever (\textbf{Qwen3-Embedding-8B}) to produce 882 candidate queries (Table~\ref{tab:pipeline-stats}).

\subsubsection{Stage III: Collision Testing and LLM Verification}
\noindent Stage~III performs adversarial quality control to (i) remove inconsistent samples and (ii) collect validated distractors. For each candidate query, we retrieve the top-$k$ nearest \emph{non-reference} documents under dense embedding similarity from the 326M-token MemoryBank using \textbf{Qwen3-Embedding-8B} (excluding documents in \textbf{RefList}). An LLM then verifies each retrieved candidate against \textbf{(Query, Answer, RefList)} and assigns one of three outcomes:
\begin{itemize}[leftmargin=*, itemsep=1pt, parsep=0pt]
    \item \textbf{Conflict:} the candidate contradicts the query/answer/reference content; we discard the entire sample.
    \item \textbf{Hard Negative:} semantically similar but does not support the answer; we retain it as an adversarial distractor.
    \item \textbf{False Negative:} provides additional valid evidence; we add it to RefList.
\end{itemize}

\subsection{Evaluation Scenarios and Input Settings}
\label{sec:eval-scenarios}

To keep the main text focused on results, we summarize the input settings used in Section~\ref{sec:experiments} and Table~\ref{tab:qa_results} here. All QA scores are produced by the same LLM-as-a-Judge protocol (Appendix~\ref{sec:eval-details}).

\textbf{Window-aware full-context evaluation.} For a model with maximum context window $N$, we only run full-context experiments on ladder scales $S$ whose \emph{tokenized} input (prompt + query + context) fits within $N$.\footnote{The ladder is constructed with fixed token budgets and prompt overhead reserved (Section~\ref{sec:corpus-ladder}), but we still enforce the constraint using the model's tokenizer for safety.}

We report one QA input setting at each scale $S$:
\begin{itemize}[leftmargin=*, itemsep=1pt, parsep=0pt]
    \item \textbf{GLOBAL:} provide the full reference corpus $\mathcal{C}^{S}$ as the model's context (reported only when the full input fits within $N$; up to 1M in Table~\ref{tab:qa_results}).
\end{itemize}

\subsection{Document-ID Evidence Interface and Localization Protocol}
\label{sec:docid-protocol}
To support architecture-agnostic evaluation of evidence access, EMB-S uses a shared \textbf{document-ID} evidence space. Each document in the MemoryBank is assigned a unique integer ID, and each ladder corpus $\mathcal{C}^S$ is a subset of these ID-labeled documents. Retrieval systems therefore return document IDs directly, and are scored by exact ID matching (Section~\ref{sec:eval-details}).

\textbf{Localization (evidence access).} Given a query and a searchable corpus $\mathcal{C}^S$, a system outputs the top-$K$ document IDs it considers relevant. For native long-context LLMs, we include explicit document-ID headers when serializing a corpus into a single input (e.g., \texttt{[DocID=123]} preceding each document) and apply a simple output canonicalization: we extract integer IDs from the model output in order and keep the first $K$ unique IDs. This reduces formatting variance while preserving the underlying evidence ranking signal.

\subsection{Raw Data Sources and Characteristics}
\label{sec:appendix-sources}

\begin{table}[htbp]
    \centering
\scriptsize
\setlength{\tabcolsep}{2pt}
\begin{tabular}{lcccc}
\toprule
\textbf{Source} & \textbf{Scale} & \textbf{Task} & \textbf{Scope} & \textbf{Focus} \\
\midrule
LongBench \citeyearpar{bai2023longbench} & 2M & Retr. & LLM & Summ. \\
L-Eval \citeyearpar{an2024leval} & 200K & QA & LLM & Summ. \\
InfiniteBench \citeyearpar{zhang2024infinitebench} & 2M & QA & LLM & Compl. \\
BABILong \citeyearpar{kuratov2024babilong} & 10M & QA & LLM & Reas. \\
Loong \citeyearpar{zhang2024loong} & 200K & QA & LLM & Reas. \\
NeedleBench \citeyearpar{xiong2023llms} & 10K & Retr. & LLM & Trace \\
RULER \citeyearpar{hsieh2024ruler} & 128K & QA & LLM & Trace \\
RAGBench \citeyearpar{friel2024ragbench} & 128K & Both & RAG & M-hop \\
LV-Eval \citeyearpar{yuan2024lveval} & 256K & QA & LLM & Bal. \\
\textbf{EMB-S (Ours)} & \textbf{326M} & \textbf{Both} & \textbf{All} & \textbf{All} \\
\bottomrule
\end{tabular}
\caption{Raw data sources for EverMemBench-S. \textbf{Scope}: target system (LLM/RAG/Memory). Sources span 10K--10M tokens.}
\label{tab:data-sources}
\end{table}

\subsection{Retrieval Metric Calculation}
Retrieval performance is evaluated using exact document identifier matching. For a given query $q$, let $G_q = \{d_1, d_2, ..., d_m\}$ be the set of ground-truth reference document IDs (single-source queries have $|G_q|=1$; multi-source queries have $|G_q|>1$), and $R_q@K = \{r_1, r_2, ..., r_k\}$ be the set of top-$K$ document IDs retrieved by the model. We report the \textbf{standard recall} (denoted as \textbf{SR@K} in Table~\ref{tab:retrieval_results}):

\begin{equation}
\text{SR@K} = \frac{|R_q@K \cap G_q|}{|G_q|}
\end{equation}

For completeness, we also report \textbf{R@1} on the single-source subset, where the metric reduces to a binary indicator of whether the unique gold document is ranked at position 1.

We further report \textbf{FR@K} on the multi-source subset, which requires retrieving \emph{all} reference documents:
\begin{equation}
\text{FR@K} = \mathbb{1}\!\left[G_q \subseteq R_q@K\right].
\end{equation}
We report FR@K averaged over queries with $|G_q|>1$.

This strict ID matching ensures that the model retrieves the exact evidence required for reasoning, rather than merely semantically similar documents which may be hard negatives.

\subsection{LLM-as-a-Judge Prompt Template}
For the Generative QA evaluation, we employ the following prompt template for LLM-as-a-Judge scoring. The judge is instructed to output \textbf{a single integer score} in $\{0,\dots,5\}$ (Figure~\ref{fig:judge-prompt}).

\input{latex/tab/prompt_judge}

\subsection{Judge Model and Human Sanity Check}
\textbf{Judge model.} Unless otherwise noted, all LLM-as-a-Judge scores in Table~\ref{tab:qa_results} are produced by Grok-4 using the prompt in Figure~\ref{fig:judge-prompt}. The judge input contains only the \{query, true answer, predicted answer\} triplet (no system identifiers) to reduce potential preference biases.

\textbf{Human sanity check.} To complement automated scoring, we additionally perform a small-scale human review on randomly sampled QAR instances, rating completeness and correctness with a 10-point rubric as a sanity check.

%% file: latex/tab/prompt_judge.tex
\begin{figure}[!htbp]
    \centering
    \begin{tcolorbox}[
        width=\columnwidth,
        colback=gray!10,
        colframe=black!50,
        title=LLM-as-a-Judge Scoring Prompt (0--5),
        enhanced,
        before skip=0pt,
        after skip=0pt,
        left=1mm,right=1mm,top=1mm,bottom=1mm
    ]
    \footnotesize
    \textbf{Instruction:} Based on the accuracy, completeness, and relevance of the predicted answer to the real answer in the context of the \textbf{query}, assign an objective score from \textbf{0 to 5} (5 being the highest, 0 the lowest). The final output can only be a single number.
    
    \textbf{Scoring Criteria:}
    \begin{itemize}[leftmargin=*]
        \item \textbf{5:} The predicted answer is exactly the same as the real answer and correctly answers the query. Differences in wording do not affect factual accuracy.
        \item \textbf{4:} The predicted answer contains all the core information of the real answer, with no errors, but includes a small amount of non-critical redundant content.
        \item \textbf{3:} The predicted answer captures the core information but differs from the real answer in some aspects. The predicted answer is slightly incomplete or imprecise, but contains no errors.
        \item \textbf{2:} The predicted answer is partially relevant to the real answer but omits a significant amount of information or deviates from the core topic of the query.
        \item \textbf{1:} The predicted answer attempts to address the query (maintains basic relevance to the topic) but provides factually incorrect information. It does not contradict the core claim of the real answer, but shows incomplete or inaccurate understanding of the topic.
        \item \textbf{0:} The predicted answer is completely unrelated to the query, consists of gibberish, or is a pure hallucination that shares no logical connection with the real answer.
    \end{itemize}
    
    \textbf{Query:} \{query\}
    
    \textbf{True Answer:} \{reference\_answer\}
    
    \textbf{Predicted Answer:} \{generated\_answer\}
    
    \textbf{Output only a single number (0, 1, 2, 3, 4, or 5):}
    \end{tcolorbox}
    \caption{\textbf{LLM-as-a-Judge scoring prompt.} The prompt used to score generated answers with a single integer score in \{0,\dots,5\}.}
    \label{fig:judge-prompt}
    \end{figure}